\documentclass[runningheads]{llncs}
\usepackage{caption} 
\newcommand{\bul}{{\small$\color{blue!80!black}\blacktriangleright~$}}

\newcommand{\exbul}[1]{\smallskip \noindent \bul \noindent\textbf{\small{{\sffamily #1.}}}}

\newcommand{\myurl}[1]{{\scriptsize \color{blue!80!black} \url{#1}}}
\newcommand{\myurlbig}[1]{{\color{blue!80!black} \url{#1}}}

\newcommand{\myalign}[2]{\text{#1} &: #2 \\}

\newcommand{\result}{\noindent\emph{The result is}:}

\newcommand{\mynot}{\text{not}~}

\newcommand{\ST}{{\small $\mathcal{ST}~$}}

\newcommand{\STbold}{{\small $\mathcal{ST}~$}}

\newcommand{\bulletHead}[1]{\smallskip \noindent\textbf{#1.}}

\newcommand{\smallBulletHead}[1]{\noindent\textbf{\small{{#1.}}}}

\newcommand{\predThF}[1]{{\operatorname{\mathsf{#1}}}}

\usepackage[sf,SF]{subfigure}
\usepackage{xcolor}
\usepackage{xspace}
\usepackage{amsmath}
\usepackage{amssymb}

\usepackage{lipsum}
\usepackage{wrapfig}
\usepackage{graphicx}

\usepackage{xcolor,colortbl}

\usepackage[sort,numbers]{natbib}

\usepackage{times}

\usepackage[polish,english]{babel}
\usepackage[T1,nomathsymbols]{polski}

\newcommand{\QS}{$\mathcal{QS}$\xspace}

\newcommand{\Pred}[1]{{\mathsf{\lowercase{#1}}}}

\usepackage[pagebackref=true,breaklinks=true,letterpaper=true,colorlinks,bookmarks=false]{hyperref}

\hypersetup{
    pdftitle={TODO}, % title
    pdfauthor={TODO},     	% author
    pdfsubject={TODO}, % subject of the document
    pdfkeywords={TODO} {TODO} {TODO}, 					% list of keywords	eator of the document
    unicode=false,          									% non-Latin characters in AcrobatÕs bookmarks
    pdftoolbar=false,        									% show AcrobatÕs toolbar?
    pdfmenubar=true,        								% show AcrobatÕs menu?
    pdffitwindow=false,     									% window fit to page when opened
    pdfstartview={FitH},    								% fits the width of the page to the window
    pdfnewwindow=true,      								% links in new window
    bookmarks=true,         								% show bookmarks bar?    
    colorlinks=true,       									% false: boxed links; true: colored links
    linkcolor=red!70!black,          							% color of internal links (change box color with linkbordercolor)
    citecolor=blue!70!black,        							% color of links to bibliography
    filecolor=blue,      									% color of file links
    urlcolor=blue!80!black          								% color of external links
}

\begin{document}
\title{{\sffamily Answer Set Programming Modulo `Space-Time'}}
\author{{\sffamily\small Carl Schultz$^1$ \and Mehul Bhatt$^2$$^,$$^3$ \and Jakob Suchan$^3$ \and Przemys{\l}aw Wa{\l}\k{e}ga$^4$}}
\authorrunning{{\sffamily C. Schultz, M. Bhatt, J. Suchan, P. Wa{\l}\k{e}ga}}


%%The DesignSpace Group. \href{www.design-space.org}{www.design-space.org}\\[5pt]

\institute{\sffamily Spatial Reasoning. \href{www.spatial-reasoning.com}{www.spatial-reasoning.com}\\[5pt]
$^1$ Aarhus University, Denmark, $^2$ \"{O}rebro University, Sweden\\$^3$ University of Bremen, Germany, $^4$University of Warsaw, Poland
}

%\institute{Princeton University, Princeton NJ 08544, USA \and
%Springer Heidelberg, Tiergartenstr. 17, 69121 Heidelberg, Germany
\maketitle              % typeset the header of the contribution
\begin{abstract}
{We present ASP Modulo `Space-Time', a declarative representational and computational framework to perform commonsense reasoning about regions with both spatial and temporal components. Supported are capabilities for mixed qualitative-quantitative reasoning, consistency checking, and inferring compositions of space-time relations; these capabilities combine and synergise for applications in a range of AI application areas where the processing and interpretation of spatio-temporal data is crucial. The framework and resulting system is the only general KR-based method for declaratively reasoning about the dynamics of `space-time' regions as first-class objects. We present an empirical evaluation (with scalability and robustness results), and include diverse application examples involving interpretation and control tasks.}
\end{abstract}

\section{\uppercase{Introduction}}
Answer Set Programming (ASP) has emerged as a robust declarative problem solving methodology with tremendous application potential \cite{gelfond1988stable,gelfond2008answer,Brewka:2011:ASP,DBLP:conf/aaai/SuchanBWS18}. Most recently, there has been heightened interest to extend ASP in order to handle specialised domains and application-specific knowledge representation and reasoning (KR) capabilities. For instance, ASP Modulo Theories (\textsc{ASPMT}) go beyond the propositional setting of standard answer set programs by the integration of ASP with Satisfiability Modulo Theories (SMT)  thereby facilitating reasoning about continuous domains \citep{joohyung-aspmt-2013,bartholomew2014system,gelfond2008answer}; using this approach, integrating knowledge sources of \emph{heterogeneous semantics} (e.g., infinite domains) becomes possible. Similarly, \textsc{Clingcon} \cite{geossc09a} combines ASP with specialised constraint solvers supporting non-linear finite integers. Other most recent extensions include the ASPMT founded \emph{non-monotonic spatial reasoning} extensions in \textsc{ASPMT(QS)} \cite{ASPMTQS-LPNMR-2015}; ASP modulo \emph{acyclicity}  \cite{ASPACyc};  \emph{probabilistic} extensions to ASP \cite{uncertainty-asp-Joohyung15}. 

Indeed, being rooted in KR, in particular non-monotonic reasoning, ASP can theoretically characterise ---and promises to serve in practice as--- a modern foundational language for several domain-specific AI formalisms, and offer a uniform computational platform for solving many of the classical AI problems involving planning, explanation, diagnosis, design, decision-making, control \citep{Brewka:2011:ASP,DBLP:conf/aaai/SuchanBWS18,DBLP:conf/date/NeubauerWSH18}. In this line of research, this paper presents ASP Modulo `{Space-Time}', a specialised formalism and computational backbone enabling generalised commonsense reasoning about `\emph{space-time objects}' and their spatio-temporal dynamics directly within the answer set programming paradigm.

\begin{figure*}[t]
\centering
\includegraphics[width=0.9\textwidth]{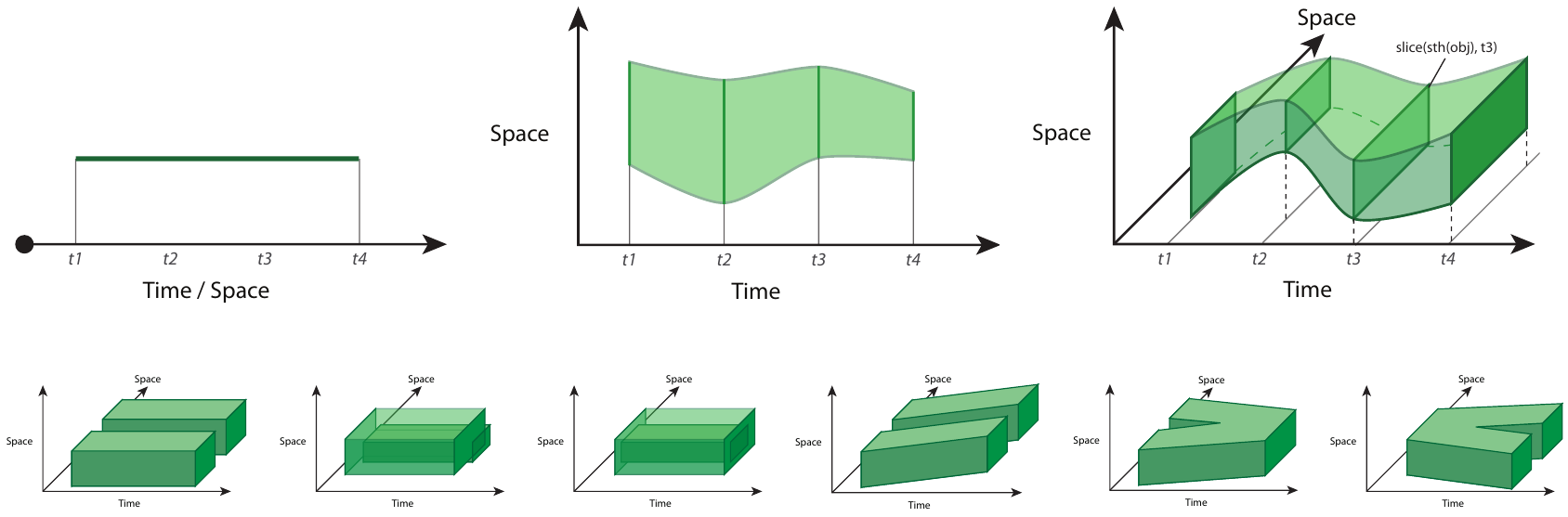}
\caption{{\sffamily\small Space-Time Histories in 1D and 2D; Spatio-temporal patterns and events, i.e. $\mathsf{discrete, overlapping, inside, parallel~movement, merge} $, and $\mathsf{split}$}.}
\label{fig:s-t-entities}
\end{figure*}

\medskip

\noindent\textbf{Reasoning about `Space-Time' (Motion)}\quad Imagine a moving object within 3D space. Here, the complete trajectory of motion of the moving object within a space-time localisation framework constitutes a 4D space-time history consisting of both spatial and temporal components -- i.e., it is a region in \emph{space-time} (Fig. \ref{fig:s-t-entities}). Regions in \emph{space}, \emph{time}, and \emph{space-time} have been an object of study across a range of disciplines such as ontology, cognitive linguistics, conceptual modeling, KR (particularly qualitative spatial reasoning), and spatial cognition and computation. Spatial knowledge representation and reasoning can be classified into two groups: topological and positional calculi \cite{aiello2007handbook,ligozat-book}. With topological calculi such as the {{Region Connection Calculus}} (RCC) \cite{randell1992spatial}, the primitive entities are spatially extended regions of space, and could be arbitrarily (but uniformly) dimensioned space-time histories. For the case of `space-time' representations,  the main focus  in the state of the art has been on axiom systems (and the study of properties resulting therefrom) aimed at pure \emph{qualitative reasoning}. In particular, axiomatic characterisations of mereotopologically founded theories with spatio-temporal regions as primitive entities are very well-studied \citep{muller1998qualitative,hazarika2005qualitative}. Furthermore, the dominant method and focus within the field of spatial representation and reasoning ---be it for topological or positional calculi--- has been primarily on relational-algebraically  founded semantics \citep{ligozat-book} in the absence of (or by discarding available) quantitative information. Pure qualitative spatial reasoning is very valuable, but it is often counterintuitive to not utilise  or discard  quantitative data if it is available (numerical information is typically available in domains involving sensing, interaction, interpretation, and control). 

%(Fig. \ref{fig:rcc8})

\noindent\textbf{Answer Set Modulo `Space-Time'}\quad Within the state of the art, it is not possible for AI applications (e.g., involving reasoning about moving objects in a vision system, control in robotic manipulation) to directly exploit commonsense representation and reasoning with `space-time' objects and their mutual spatial-temporal relationships as first-class entities within a robust KR framework such as ASP. The main contributions of the paper are:\quad {\small\textbf{(1)}}. Developing a systematic formal account and associated compuatational characterisation of a `space-time' theory as a general language founded in answer set programming; the focus is on declarative modelling, commonsense inference and question-answering with space-time objects and their mutual relationships as first-class objects;\quad  {\small\textbf{(2)}}. Support of mixed qualitative-quantitative reasoning and dynamic \emph{quantification} (i.e., grounding of real world parameters); this is very powerful, e.g., when only partial information is available, (sensor) data is noisy, or when \emph{quantification} --- is not needed or can be delayed;\quad  {\small\textbf{(3)}}. Demonstrating, by running examples and an empirical evaluation, the applicability of the resulting general reasoning system to support reasoning about space-time histories in diverse application scenarios focussing on interpretation and control. The proposed model is implemented using \textsc{clingo} \citep{gebser2011potassco,gekakasc14b}; to the best of our knowledge, no systematic realisation of a general declarative method supporting native space-time histories and relationships therof currently exists (be it mixed qualitative-quantitative reasoning, or even purely qualitative reasoning).\footnote{Implementation and examples may be consulted here: \href{http://think-spatial.org/ASP-ST.zip}{http://think-spatial.org/ASP-ST.zip}.}

\section{\uppercase{ASP Modulo `Space-Time'}}\label{sec:st-domain}

\subsection{Space-Time Histories}\label{sec:aspmt-lang}

The spatio-temporal domain (\STbold)  that we focus on in our formal framework consists of the following ontology:

\medskip

\bulletHead{Spatial Domains} Spatial domain entities include \emph{points} and \emph{simple polygons}:  a $2D$ \emph{point} is a pair of reals $x,y$;  a \emph{simple polygon} $P$ is defined by a list of $n$ vertices (points) $p_0, \dots, p_{n-1}$ such that the boundary is non-self-intersecting, i.e., no two edges of the polygon intersect. We denote the number of vertices in $P$ with $|P|$. A polygon is \emph{ground} if all vertices are assigned real values. A translation vector $t$ is a pair of reals $t_x, t_y$. Given point $p = (x,y)$ and translation vector $t$ then $p + t = (x+t_x, y+t_y)$. A \emph{translation} is a ternary relation between two polygons $P,Q$ and a translation vector $t$ such that: $|P| = |Q| = n$ and $p_i = q_i + t$ where $p_i$ is the $i^\text{th}$ vertex in $P$ and $q_i$ is the $i^\text{th}$ vertex in $Q$, for $0 \leq i < n$ . A translation vector $t$ is \emph{ground} if $t_x,t_y$ are assigned real values, otherwise it is \emph{unground}.\footnote{For brevity we focus on $2$D spatial entities; our approach also readily extends to $3$D spatial entities, and in general $n$D points, polytopes, and translation vectors.}

%\bulletHead{Spatial Domains} Domain entities in \QS include \emph{points}, \emph{line segments}, and \emph{simple polygons}:  a $2D$ \emph{point} is a pair of reals $x,y$;  a \emph{line segment} is a pair of end points $p_1, p_2$ ($p_1 \neq p_2$);  a \emph{simple polygon} $P$ is defined by a list of $n$ vertices (points) $p_0, \dots, p_{n-1}$ such that the boundary is non-self-intersecting, i.e., no two edges of the polygon intersect. We denote the number of vertices in $P$ with $|P|$. A polygon is \emph{ground} if all vertices are assigned real values.

\bulletHead{Temporal domain $\mathcal{T}$} The temporal dimension is constituted by an infinite set of time points -- each time point is a real number. The time-line is given by a linear ordering $<$ of time-points.

\bulletHead{\STbold Histories} Consider a moving two-dimensional spatial object $s$, e.g. represented by a polygon at each time point. If we treat time as an additional dimension, then we can represent $s$ as a three-dimensional object in space-time. Intuitively, at each time point, the corresponding space-time region of $s$ has a 2D spatial representation (a spatial \emph{slice}). The space-time object is formed by taking all such slices over time.

\smallskip

An $\mathcal{ST}$ \emph{object} $o \in O$ is a variable associated with an ST domain $D$ (e.g. the domain of $2D$ polygons over time). An \emph{instance} of an object $i \in D$ is an element from the domain. Given $O = \{o_1, \dots, o_n\}$, and domains $D_1, \dots, D_n$ such that $o_i$ is associated with domain $D_i$, then a \emph{configuration} of objects $\psi$ is a one-to-one mapping between object variables and instances from the domain, $\psi(o_i) \in D_i$. For example, a variable $o_1$ is associated with the domain $D_1$ of moving 2D points over time. An \STbold point moving in a straight line starting at spatial coordinates $(0,0)$ at time $0$ and arriving at 2D spatial coordinates $(10,0)$ at time $1$ is an instance of $D_1$. A configuration is defined that maps $o_1$ to a 3D line with end points $(0,0,0), (10,0,1)$ i.e. $\psi(o_1) = [(0,0,0), (10,0,1)]$.

%\bulletHead{\STbold\ Relations} Let $D_1, \dots, D_n$ be spatio-temporal domains. A spatio-temporal relation $r$ of arity $n$ ($0 < n$) is defined as $ r \subseteq D_1 \times \dots \times D_n $. That is, each spatio-temporal relation is an equivalence class of instances of \STbold\ objects. Given a set of objects $O$, a relation $r$ of arity $n$ can be asserted as a constraint that must hold between objects $o_1, \dots, o_n \in O$, denoted $r_{1, \dots, n}$. The constraint $r_{1, \dots, n}$ is satisfied by configuration $\psi$ if $\big(\psi(o_1), \dots, \psi(o_n) \big) \in r$. For example, if $pp$ is a topological relation \emph{proper part}, and $O=\{o_1, o_2\}$ is a set of moving polygon objects, then $pp_{1,2}$ is the constraint that moving polygon $o_1$ is a proper part of $o_2$.

\bulletHead{\STbold\ Relations} Let $D_1, \dots, D_n$ be spatio-temporal domains. A spatio-temporal relation $r$ of arity $n$ ($0 < n$) is defined as $ r \subseteq D_1 \times \dots \times D_n $. That is, each spatio-temporal relation is an equivalence class of instances of \STbold\ objects. Given a set of objects $O$, a relation $r$ of arity $n$ can be asserted as a constraint that must hold between objects $o_1, \dots, o_n \in O$, denoted $r(o_1, \dots, o_n)$. The constraint $r(o_1, \dots, o_n)$ is satisfied by configuration $\psi$ if $\big(\psi(o_1), \dots, \psi(o_n) \big) \in r$. For example, if $pp$ is a topological relation \emph{proper part}, and $O=\{o_1, o_2\}$ is a set of moving polygon objects, then $pp(o_1, o_2)$ is the constraint that moving polygon $o_1$ is a proper part of $o_2$.

Table~\ref{tab:rels} presents definitions for \ST relations that hold between $s_1$ and $s_2$, where $t, t^\prime$ range over a (dense) time interval with start and end time points $t_{0}$ and $t_{N}$ in which $s_1$ and $s_2$ occur and $t \leq t^\prime$. We define mereotopological relations using the Region Connection Calculus (RCC) \cite{randell1992spatial}: all spatio-temporal RCC relations between \ST regions are defined based on the RCC relations of their slices (for simplicity we use the same names for spatial and spatio-temporal RCC relations). \ST regions \emph{split} (conversely, \emph{merge}) if their spatial slices are initially parts and end up disconnected. \ST region $s$ \emph{grows} or \emph{shrinks} if the area monotonically increases or decreases, respectively. \ST region $s$ \emph{moves} if the centre point changes, and region $s_1$ moves \emph{away} from, \emph{towards} $s_2$ if the centre point distance ($\Delta$) increases, decreases, and \emph{parallel} if the vector between centre points does not change. An \ST region $s_1$ \emph{follows} \ST region $s_2$ if, at each time step, $s_1$ moves towards a previous location of $s_2$, and $s_2$ moves away from a previous location of $s_1$; we introduce a user-specified maximum duration threshold $\alpha$ between these two time points to prevent unwanted scenarios being defined as \emph{follows} events such as $s_1$ taking one step towards $s_2$ and then stopping while $s_2$ continues to move away from $s_1$.

%\begin{table}[tp]
\begin{table}[t]
\centering
\scriptsize
%\begin{tabular}{|p{30 ex} | l |}
\begin{tabular}{| l | l |}
\hline
\textbf{\scriptsize\sffamily Relation} & \textbf{\scriptsize\sffamily Definition} \\
\hline
\textbf{\scriptsize\sffamily Topology} & \\
\hline
disconnects (DC) & $\forall t ~ \Pred{dc}(s_1(t), s_2(t))$ \\
discrete from (DR) & $\forall t ~ \Pred{dr}(s_1(t), s_2(t))$ \\
part of (P) & $\forall t ~ \Pred{p}(s_1(t), s_2(t))$ \\
%non-tangential proper part (NTPP) & $\forall t ~ \Pred{ntpp}(s_1(t), s_2(t))$ \\
non-tangential & $\forall t ~ \Pred{ntpp}(s_1(t), s_2(t))$ \\
proper part (NTPP) & \\
equal (EQ) & $\forall t ~ \Pred{eq}(s_1(t), s_2(t))$ \\

contacts (C) & $\exists t ~ \Pred{c}(s_1(t), s_2(t))$ \\
overlaps (O) & $\exists t ~ \Pred{o}(s_1(t), s_2(t))$ \\
partially overlaps (PO) & $\exists t ~ \Pred{po}(s_1(t), s_2(t))$ \\
externally connects (EC) & $\Pred{dr}(s_1,s_2) \wedge \exists t ~ \Pred{ec}(s_1(t), s_2(t))$ \\
proper part (PP) & $\Pred{p}(s_1,s_2) \wedge \exists t ~ \Pred{pp}(s_1(t), s_2(t))$ \\
tangential proper part (TPP) & $\Pred{p}(s_1,s_2) \wedge \exists t ~ \Pred{tpp}(s_1(t), s_2(t))$ \\

%split & $\exists t \exists t^\prime \forall t_i \forall t_j \forall t_k \Pred{p}(s_1(t_i),s_2(t_i)) \wedge \Pred{po}(s_1(t_j),s_2(t_j))$ \\
%& $\wedge \Pred{dr}(s_1(t_k),s_2(t_k))$, such that $t_i \leq t \leq t_j \leq t^\prime \leq t_k$\\
%& $\wedge \Pred{dr}(s_1(t_k),s_2(t_k))$, such that $t_i \leq t \leq t_j \leq t^\prime \leq t_k$\\
%split & $\forall t \forall t^\prime \forall t^{\prime\prime} \Pred{p}(s_1(t),s_2(t)) \wedge \Pred{po}(s_1(t^\prime),s_2(t^\prime)) \wedge \Pred{dr}(s_1(t^{\prime}),s_2(t^{\prime}))$ \\
split & $\Pred{p}(s_1(t_{0}), s_2(t_{0})) \wedge \Pred{dc}(s_1(t_{N}), s_2(t_{N}))$ \\
merge & $\Pred{dc}(s_1(t_{0}), s_2(t_{0})) \wedge \Pred{p}(s_1(t_{N}), s_2(t_{N}))$ \\

\hline
\textbf{\scriptsize\sffamily Size} & \\
\hline

fixed size & $\forall t \forall t^\prime \big( \Pred{area}(s(t)) = \Pred{area}(s(t^\prime)) \big)$ \\
\hline
grows & $\neg \Pred{fixed\_size}(s) \wedge$ \\
  & $\forall t \forall t^\prime \big(\Pred{area}(s(t)) \leq \Pred{area}(s(t^\prime)) \big)$ \\
\hline
shrinks & $\Pred{reverse}(\Pred{grows}(s_1, s_2))$\\
\hline
\textbf{\scriptsize\sffamily Movement} & \\
\hline
moves & $\exists t \exists t^\prime p(t) \neq p(t^\prime)$ \\
\hline
move parallel & $\Pred{moves}(s_1) \wedge$ \\
%& $\forall t \forall t^\prime \Pred{\Delta}(p_1(t),p_2(t)) = \Pred{\Delta}(p_1(t^\prime),p_2(t^\prime))$ \\
& $\forall t \forall t^\prime \big( p_2(t) - p_1(t)\big)=\big(p_2(t^\prime) - p_1(t^\prime)\big)$ \\
\hline
towards & $\Pred{moves}(s_1) \wedge \neg \Pred{moves\_parallel}(s_1,s_2) \wedge$ \\
& $\forall t \forall t^\prime \Pred{\Delta}(p_1(t),p_2(t)) \geq \Pred{\Delta}(p_1(t^\prime),p_2(t^\prime))$ \\
\hline
away & $\Pred{reverse}(\Pred{towards}(s_1, s_2))$ \\
\hline
follows & $\forall t^\prime \exists t ~ \Pred{duration(t,t^\prime) \leq \alpha} $ \\
& $\wedge \Pred{\Delta}(p_1(t),p_2(t)) > \Pred{\Delta}(p_1(t^\prime),p_2(t))$ \\
& $\wedge \Pred{\Delta}(p_1(t),p_2(t)) < \Pred{\Delta}(p_1(t),p_2(t^\prime))$ \\
\hline
\end{tabular}
\caption{{\footnotesize\sffamily Relations between \ST regions $s_1, s_2$ over time interval $I = [t_{0}, t_{N}]$; $t, t^\prime$ range over $I$ with $t \leq t^\prime$; $\Pred{reverse}(R)$ denotes the definition of relation $R$ with reversed temporal ordering, $t^\prime \leq t$; $p_i(t_j)$ is the centre point of $s_i$ at $t_j$; $\Delta$ is the Euclidean distance between two points; $\alpha$ is a user-specified temporal threshold.}}
\label{tab:rels}
\end{table}

\subsection{Space-Time Semantics as Polynomial Constraints}\label{sec:polysolve}
One approach for formalising the semantics of spatial reasoning is by encoding qualitative spatial relations as systems of polynomial equations and inequalities \cite{bhatt2011clp,ASPMTQS-LPNMR-2015}. The task of determining whether a set of spatial relations is consistent is then equivalent to determining whether the set of polynomial constraints are satisfiable. Given a system of polynomial constraints over real variables $X$, the constraints are satisfiable if there exists some real value for each variable in $X$ such that all the polynomial constraints are simultaneously satisfied.\footnote{The worst case complexity of solving a system of non-linear polynomial constraints over $n$ real variables is $O(2^{2^n})$ \cite{arnon1984cylindrical} owing to the Cylindrical Algebraic Decomposition algorithm \cite{Collins1991299}, which is implemented in the solver z3 \citep{de2008z3}. Although not relevant to this paper, it is worth pointing out that we use a (sound and complete) polynomial constraint solver that determines whether a system of non-linear polynomial constraints is satisfiable, based on an integration of Satisfiability Modulo Theories solver z3 \cite{de2008z3} and numerical optimisation \cite{schultz2016numerical} with the library NLopt \cite{johnson2014nlopt} using BOBYQA \cite{powell2009bobyqa}. The employed polynominal encodings are highly optimised (e.g., by symmetry-based pruning heuristics \citep{DBLP:conf/cosit/SchultzB15}) for the specific spatio-temporal context.} For example, let point $p$ be defined by real coordinates $x_p,y_p$, and let circle $c$ be defined by the centre point $x_c,y_c$ and real radius $r_c$. A point $p$ is incident to the interior of a circle $c$ if the distance between $p$ and the centre of $c$ is less than the radius of $c$: $(x_p - x_c)^2 + (y_p - y_c)^2 < r_c^2$.  If there exists an assignment of real values to the variables (e.g., $x_p = 3.5, x_c = 10.5$, etc.) that satisfies all polynomial constraints, then the qualitative spatial relations are consistent. Continuing with the example, if we now add the relation that point $p$ is also incident to the boundary of $c$: $(x_p - x_c)^2 + (y_p - y_c)^2 = r_c^2$
\noindent and we reformulate the system of constraints we get: $(d_{pc} < r_c) \wedge (d_{pc} = r_c)$. Distance $d_{pc}$ cannot be both less than and equal to the radius $r_c$, and thus the system of polynomial constraints is inconsistent, and no configuration of points and circles (within Euclidean space) exists that can satisfy this set of qualitative spatial relations. 

\smallskip

%\quad A \emph{term} is either a variable, constant, or a structure $f(t_1, \dots, t_n)$ with functor $f$ applied to terms $t_1, \dots, t_n$. An \emph{atom} $p(t_1,\dots,t_n)$ is a predicate $p$ of arity $n$ with terms $t_1,\dots,t_n$. An atom is \emph{ground} if it is variable free. A literal is an atom $p$ or its default negation $\mynot p$, where literal $\mynot p$ is assumed to hold unless $p$ is derived to be true. Classic negation of atom $p$, denoted $-p$, is the complement of $p$ such that $p \wedge -p \equiv \bot$. 

%The Herbrand Base $H_P$ of ASP program $P$ is the set of ground atoms that can be made from the constants and function symbols of $P$. Given $I \subseteq H_P$, the \emph{reduct} $P^I \subseteq H_P$ is a grounding of $P$ such that it does \emph{not} include (a) any rule whose body contains a literal $\mynot c_i$ where $c_i \in I$, and (b) any remaining negative literals in the remaining rules. Set $A \subseteq H_P$ is a \emph{stable model} of $P$ iff it is a minimal model of the reduct $P^A$.

\subsection{Spatio-Temporal Consistency}
Consider the topological \emph{disconnected} relation. There is no polygon that is \emph{disconnceted} from itself, i.e. the relation is \emph{irreflexive}. Algebraic properties of \ST relations are expressed as the following ASP rules and constraints.\footnote{Standard stable model semantics is applicable \cite{ferraris2011stable}, \cite{gelfond1988stable}, and \cite{ferraris2005answer}. An ASP program $P$ consists of a finite set of universally quantified \emph{rules} of the form $h \leftarrow b_1, \dots, b_n, \mynot c_1, \dots, \mynot c_m$ such that $h$ is an atom, and the expression $b_1, \dots, b_n, \mynot c_1, \dots, \mynot c_m$ is a conjunction of atoms. ASP \emph{facts} are rules of the form $h \leftarrow \top$, and ASP \emph{constraints} are rules of the form $\bot \leftarrow b_1, \dots, b_n, \mynot c_1, \dots, \mynot c_m$.}
{\small\sffamily
\begin{equation}
\begin{aligned}
\myalign{$r$ is \emph{reflexive}}{~~\; r(A,A) \leftarrow entity(A)}
\myalign{$r$ is \emph{irreflexive}}{-r(A,A) \leftarrow entity(A)}
\myalign{$r$ is \emph{symmetric}}{~~\; r(B,A) \leftarrow r(A,B)}
\myalign{$r$ is \emph{asymmetric}}{-r(B,A) \leftarrow r(A,B)}
\myalign{$r_2$ is \emph{converse} of $r_1$}{r_2(B,A) \leftarrow r_1(A,B)}
\myalign{$r_1$ is \emph{implies} of $r_2$}{r_2(A,B) \leftarrow r_1(A,B)}
\myalign{$r_1,r_2$ are \emph{mutually inconsistent} }{\bot \leftarrow r_1(A,B), r_2(A,B)}
\myalign{$r_1,r_2,r_3$ are \emph{transitively inconsistent}}{\bot \leftarrow r_1(A,B), r_2(B,C), r_3(A,C)}
\end{aligned}
\end{equation}
}

%{\centering\small\sffamily
%\noindent $r$ is {reflexive}:\quad $r(A,A) \leftarrow entity(A)$
%
%\noindent $r$ is {irreflexive}:\quad $-r(A,A) \leftarrow entity(A)$
%
%\noindent $r$ is {symmetric}:\quad $r(B,A) \leftarrow r(A,B)$
%
%\noindent $r$ is {asymmetric}:\quad $-r(B,A) \leftarrow r(A,B)$
%
%\noindent $r_2$ is {converse} of $r_1$:\quad $r_2(B,A) \leftarrow r_1(A,B)$
%
%\noindent $r_1$ {implies} $r_2$:\quad $r_2(A,B) \leftarrow r_1(A,B)$
%
%\noindent $r_1,r_2$ are {mutually inconsistent}:\quad $\bot \leftarrow r_1(A,B), r_2(A,B)$
%
%\noindent $r_1,r_2,r_3$ are {transitively inconsistent}:\quad $\bot \leftarrow r_1(A,B), r_2(B,C), r_3(A,C)$
%
%}
%
%\smallskip

\noindent We have automatically derived these properties using our polynomial constraint solver \emph{a priori} and generated the corresponding ASP rules.  A violation of these properties corresponds to \emph{3-path inconsistency} \citep{ligozat-book}, i.e. there does not exist any combination of polygons that can violate these properties. In particular, a total of $1586$ space-time constraints result.\footnote{These may be consulted in the files ``spatial\_invariance.lp'' and ``movement\_invariance.lp'' in the submitted source code.}

\medskip

\bulletHead{Ground Polygons}\quad We can determine whether \ST relation $r$ holds between two ground polygons $P,Q$ by directly checking whether the corresponding polynomial constraints are satisfied, i.e. polynomial constraint variables are replaced by the real values assigned to the ground polygon vertices. This is accomplished during the \emph{grounding} phase of ASP. E.g. two ground polygons are \emph{disconnected} if the distance between them is greater than zero.

\bulletHead{Unground Translation}\quad Given ground polygons $P_0,P_1$, \emph{unground} polygon $P^{\prime}_0$, and unground translation $t = (t_x,t_y)$, let $P_{0}^\prime$ be a $t$ translation of $P_0$ such that $r$ holds between $P_{0}^\prime, P_1$. The (exact) set of real value pairs that can be assigned to $(t_x,t_y)$ such that $P_{0}^\prime, P_1$ satisfy $r$ is precisely determined using the Minkowski sum method \cite{wallgrun2013topological}; we refer to this set as the \emph{solution set} of $t$ for $r$. Given $n$ ground polygons $P_1, \dots, P_n$, and $n$ relations $r_1, \dots, r_n$ such that relation $r_i$ is asserted to hold between polygon $P_0, P_i$, for $1 \leq i \leq n$, let $M_i$ be the solution set of $t$ for $r_i$. The conjunction of relations $r_1, \dots, r_n$ is consistent if the \emph{intersection} of solution sets $M_1, \dots, M_n$ is non-empty. Computing and intersecting solution sets is accomplished during the \emph{grounding} phase of ASP.

\bulletHead{\STbold\ Relation Consistency} In the following tasks the input is a set of objects $O$ and a set of qualitative spatio-temporal relations $R$ between those objects: (1) \emph{Consistency.}\quad Determine whether there exists a configuration $\psi$ of $O$ that satisfies all relation constraints in $R$. Such a configuration is called a \emph{consistent configuration}; (2). \emph{Generating configurations.}\quad Return a consistent configuration $\psi$ of $O$.

\section{\uppercase{Reasoning with ASP Modulo Space-Time}}\label{sec:eval}
We have implemented our \ST reasoning module in Clingo (v5.1.0) \cite{gebser2011potassco,gekakasc14b}. Table~\ref{tab:asp-preds} presents our system's predicate interface. Our system provides special predicates for (1) declaring spatial objects, and (2) relating objects spatio-temporally. Each \ST object is represented with \emph{st\_object/3} relating the identifier of the \ST entity, time point of this slice, and identifier of the associated geometric representation.

\smallskip

%% testasp
\includegraphics[width=\textwidth]{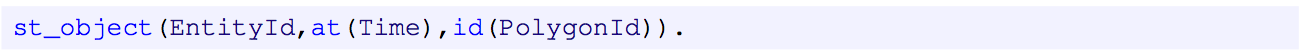}

%\scriptsize
%\begin{minted}[bgcolor=blue!5!white]{prolog}
%st_object(EntityId,at(Time),id(PolygonId)).
%\end{minted}
%\normalsize

\smallskip

Polygons are represented using the \emph{polygon/2} predicate that relates an identifier of the geometric representation with a list of $x$,$y$ vertex coordinate pairs, e.g.:

%Detected contours for each image in the dataset are represented using the \emph{polygon/2} predicate that relates an identifier of the geometric representation with a list of $x$,$y$ pairs, e.g.:

\includegraphics[width=\textwidth]{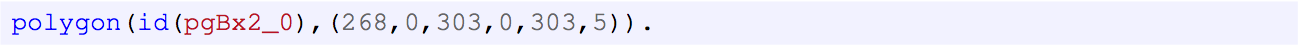}

%\scriptsize
%\begin{minted}[bgcolor=blue!5!white]{prolog}
%polygon(id(pgBx2_0),(268,0,303,0,303,5)).
%\end{minted}
%\normalsize

\begin{table}[t]
\centering
\scriptsize
\begin{tabular}{| l | p{5.5cm} |}
\hline
\textbf{\scriptsize\sffamily Predicate} & \textbf{\scriptsize\sffamily Description} \\
\hline
\textbf{\STbold Entities} & \\
\hline
$\predThF{polygon(Pg, (X1, Y2, ..., Xn, Yn))}$ & Polygon $\predThF{Pg}$ has $n$ ground vertices $(x_1, y_1), \dots, (x_n, y_n)$.\\
$\predThF{translation(Pg1, Pg2)}$ & Polygon $\predThF{Pg2}$ is an unground translation of $\predThF{Pg1}$. \\
$\predThF{st\_object(E)}$ &  $\predThF{E}$ is a spatio-temporal entity.\\
$\predThF{st\_object(E, at(Time), id(Pg))}$ & 2D polygon $\predThF{Pg}$ is a spatial \emph{slice} of spatio-temporal entity $\predThF{E}$ at time point $\predThF{Time}$.\\
\hline
\textbf{\STbold Relations} & \\
\hline
$\predThF{spacetime(STAspect, E, time(T1,T2))}$ & Derive unary ST relations for $\predThF{STAspect}$ (topology, size, or movement) for entity $\predThF{E}$ from time $\predThF{T1}$ to $\predThF{T2}$. \\
$\predThF{spacetime(STAspect, E1, E2, time(T1,T2))}$ & Derive binary ST relations for $\predThF{STAspect}$ (topology, size, or movement) between entities $\predThF{E1,E2}$ from time $\predThF{T1}$ to $\predThF{T2}$.  \\
$\predThF{topology(Rel, E1, E2,time(T1,T2))}$ & Topological relation $\predThF{Rel}$ is asserted to hold between ST entities $\predThF{E1,E2}$ from time $\predThF{T1}$ to $\predThF{T2}$.\\
$\predThF{size(Rel, E1, E2, time(T1,T2))}$ & Size relation $\predThF{Rel}$ is asserted to hold between ST entities $\predThF{E1,E2}$ from time $\predThF{T1}$ to $\predThF{T2}$. \\
$\predThF{movement(Rel, E, time(T1,T2))}$ & Unary movement relation $\predThF{Rel}$ is asserted to hold for ST entity $\predThF{E}$ from time $\predThF{T1}$ to $\predThF{T2}$. \\
$\predThF{movement(Rel, E1, E2, time(T1,T2))}$ & Binary movement relation $\predThF{Rel}$ is asserted to hold between ST entities $\predThF{E1,E2}$ from time $\predThF{T1}$ to $\predThF{T2}$. \\
$\predThF{spatial(witness, E, EWitness)}$ & Ground entity $\predThF{EWitness}$ is a consistent witness for unground entity $\predThF{E}$. \\
\hline
\end{tabular}
\caption{{\small \STbold entities and relation predicates.}}
\label{tab:asp-preds}
\end{table}

\smallskip

\smallBulletHead{Deriving \ST relations} the predicate \emph{spacetime/3} is used to specify the entities between which \ST relations should be derived:

\smallskip

\includegraphics[width=\textwidth]{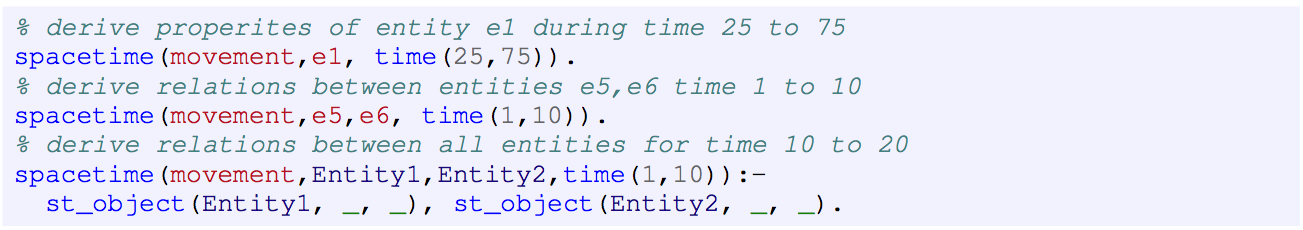}

%\scriptsize
%\begin{minted}[bgcolor=blue!5!white]{prolog}
%% derive properites of entity e1 during time 25 to 75
%spacetime(movement,e1, time(25,75)).
%% derive relations between entities e5,e6 time 1 to 10
%spacetime(movement,e5,e6, time(1,10)).
%% derive relations between all entities for time 10 to 20
%spacetime(movement,Entity1,Entity2,time(1,10)):- 
%  st_object(Entity1, _, _), st_object(Entity2, _, _).
%\end{minted}
%\normalsize

\smallskip

\smallBulletHead{Purely qualitative reasoning} if no geometric information for slices is given then our system satisfies $3$-consistency, e.g. the following program includes transitively inconsistent spatio-temporal relations:

\smallskip

\includegraphics[width=\textwidth]{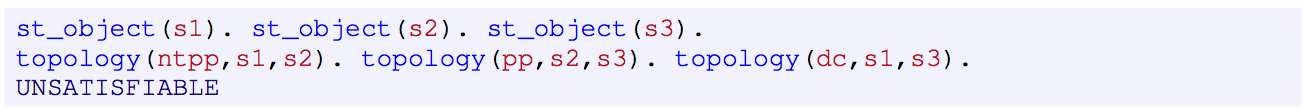}

%\scriptsize
%\begin{minted}[bgcolor=blue!5!white]{prolog}
%st_object(s1). st_object(s2). st_object(s3).
%topology(ntpp,s1,s2). topology(pp,s2,s3). topology(dc,s1,s3).
%UNSATISFIABLE
%\end{minted}
%\normalsize

\smallskip

\smallBulletHead{Mixed qualitative-numerical reasoning} a new \ST object can be specified that consists of \emph{translated} slices of a given \ST object. Our system determines whether translations exist that satisfy all given spatio-temporal constraints. Our system produces the solution set and a spatial witness that minimises the translation distance.

\smallskip

\includegraphics[width=\textwidth]{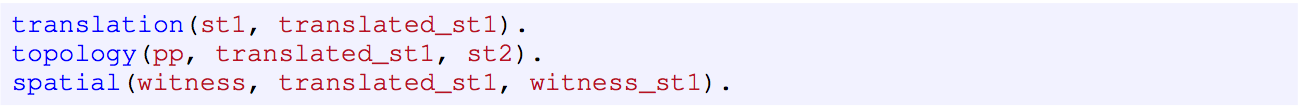}

%\scriptsize
%\begin{minted}[bgcolor=blue!5!white]{prolog}
%translation(st1, translated_st1).
%topology(pp, translated_st1, st2).
%spatial(witness, translated_st1, witness_st1).
%\end{minted}
%\normalsize

\subsection{Application Examples: Interpretation and Control}

\begin{figure}[t]
\begin{center}
\includegraphics[height=0.485\textwidth]{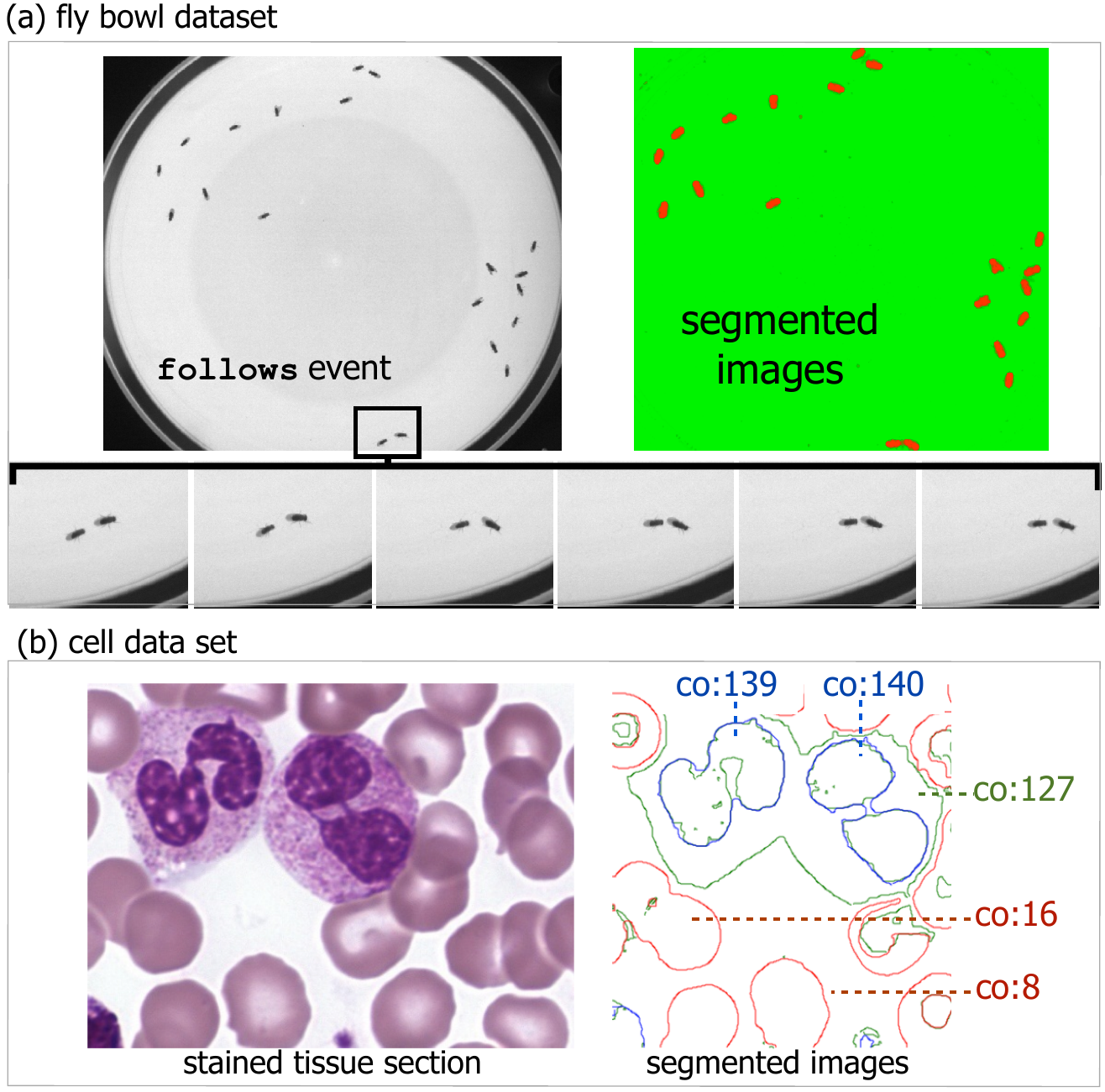}
\includegraphics[height=0.465\textwidth]{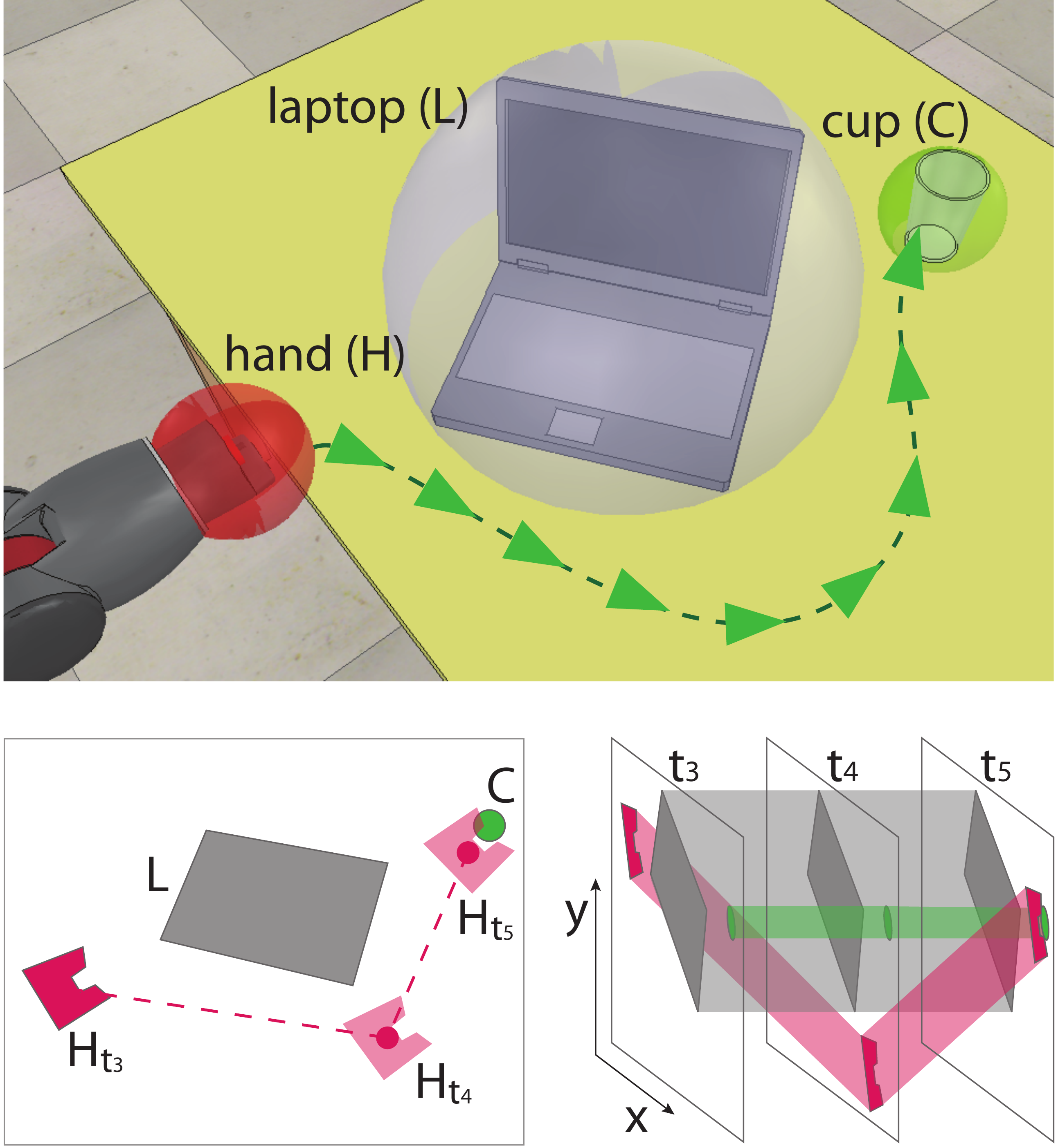} 
\end{center}
\caption{{\small\sffamily Application:\quad Insect Behaviour, Cell Biology, and Cognitive Robotics}}
\label{fig:medical}
\end{figure}

\exbul{1. INSECT BEHAVIOUR}\quad In this section we describe how spatio-temporal relations are derived from a large dataset of fly movement video data used to study the social interactions of flies.\footnote{Data provided by K. Branson from Janelia Research Campus: \myurl{https://www.janelia.org/lab/branson-lab}; accessible from the \emph{ilastik} website: \myurl{http://ilastik.org/download.html}} The dataset consists of 20 flies in a bowl, captured in 200 image frames (130 MB). Figure~\ref{fig:medical}(a) illustrates example images of the dataset and segmentation. We performed initial image segmentation and animal tracking using the \emph{ilastik} interactive toolkit \cite{sommer2011ilastik}. We then parse the output into our ASP predicates: \emph{st\_object/3} and \emph{polygon/2}.

\smallskip

\noindent\textbf{Example 1.1}.\quad  {\sffamily\small  Derive \ST movement relations between all pairs of flies for the first time step}:

\smallskip

\includegraphics[width=\textwidth]{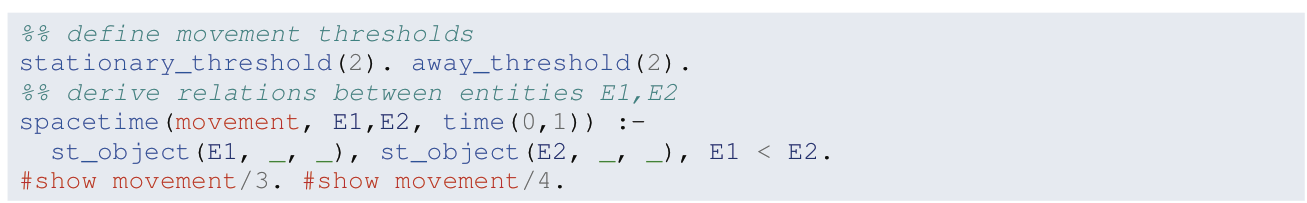}

%\scriptsize
%\begin{minted}[bgcolor=blue!5!white]{prolog}
%%% define movement thresholds
%stationary_threshold(2). away_threshold(2).
%%% derive relations between entities E1,E2
%spacetime(movement, E1,E2, time(0,1)) :-
%  st_object(E1, _, _), st_object(E2, _, _), E1 < E2.
%#show movement/3. #show movement/4.
%\end{minted}
%\normalsize

\smallskip

\result

\smallskip

\includegraphics[width=\textwidth]{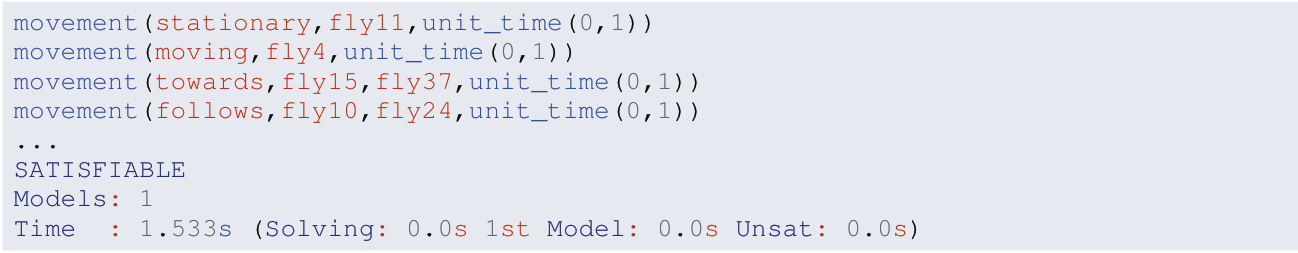}

%\scriptsize
%\begin{minted}[bgcolor=blue!5!white]{prolog}
%movement(stationary,fly11,unit_time(0,1)) 
%movement(moving,fly4,unit_time(0,1)) 
%movement(towards,fly15,fly37,unit_time(0,1)) 
%movement(follows,fly10,fly24,unit_time(0,1)) 
%...
%SATISFIABLE
%Models: 1
%Time  : 1.533s (Solving: 0.0s 1st Model: 0.0s Unsat: 0.0s)
%\end{minted}
%\normalsize

\smallskip

The extract of the results shows that, during the first time step: $fly11$ is stationary; $fly4$ is moving; $fly15$ is moving towards $fly37$; $fly10$ is following $fly24$. 

\smallskip

\noindent\textbf{Example 1.2}.\quad  {\sffamily\small Derive all spacetime movement relations between flies $fly25$, $fly24$ for the entire video}:

\includegraphics[width=\textwidth]{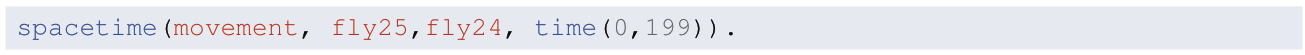}

%\scriptsize
%\begin{minted}[bgcolor=blue!5!white]{prolog}
%spacetime(movement, fly25,fly24, time(0,199)).
%\end{minted}
%\normalsize

\smallskip

\result

\smallskip

\includegraphics[width=\textwidth]{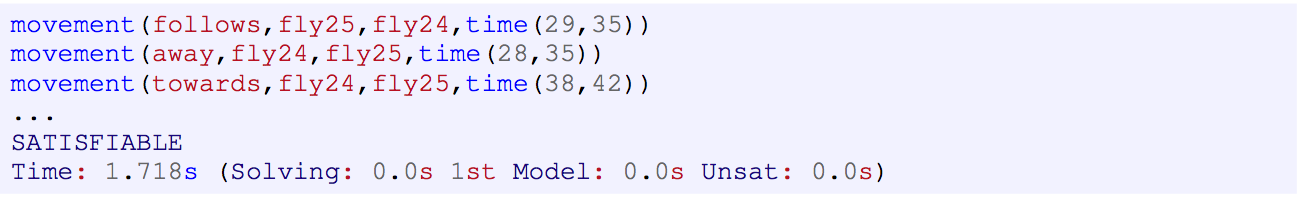}

%\scriptsize
%\begin{minted}[bgcolor=blue!5!white]{prolog}
%movement(follows,fly25,fly24,time(29,35)) 
%movement(away,fly24,fly25,time(28,35)) 
%movement(towards,fly24,fly25,time(38,42)) 
%...
%SATISFIABLE
%Time: 1.718s (Solving: 0.0s 1st Model: 0.0s Unsat: 0.0s)
%\end{minted}
%\normalsize

\smallskip

The extract of the results shows that: during time period $[29, 35]$ $fly25$ is following $fly24$; during time period $[28, 35]$ $fly24$ is moving away from $fly25$; during time period $[38, 42]$ $fly24$ is moving towards $fly25$.  

\smallskip

\noindent\textbf{Example 1.3}.\quad  {\sffamily\small Find flies that are \emph{near} each other at time $25$ and exhibit \emph{follows} behaviour for at least 3 time units during period from time $25$ to $35$}:

\smallskip

\includegraphics[width=\textwidth]{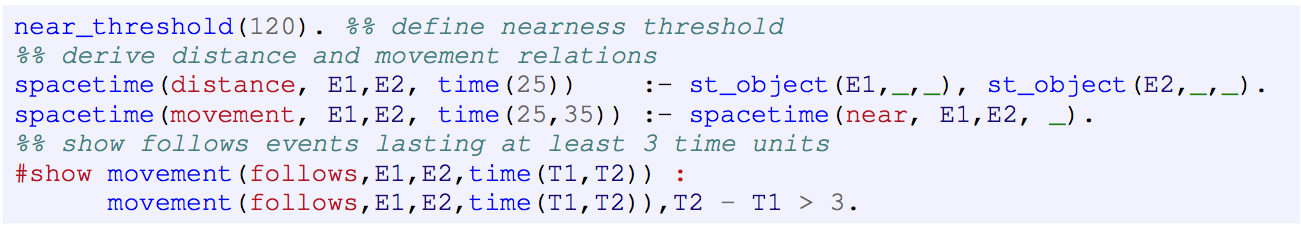}

%\scriptsize
%\begin{minted}[bgcolor=blue!5!white]{prolog}
%near_threshold(120). %% define nearness threshold
%%% derive distance and movement relations
%spacetime(distance, E1,E2, time(25))    :- st_object(E1,_,_), st_object(E2,_,_).
%spacetime(movement, E1,E2, time(25,35)) :- spacetime(near, E1,E2, _).
%%% show follows events lasting at least 3 time units
%#show movement(follows,E1,E2,time(T1,T2)) :
%      movement(follows,E1,E2,time(T1,T2)),T2 - T1 > 3.
%\end{minted}
%\normalsize

\smallskip

\result 

\smallskip

\includegraphics[width=\textwidth]{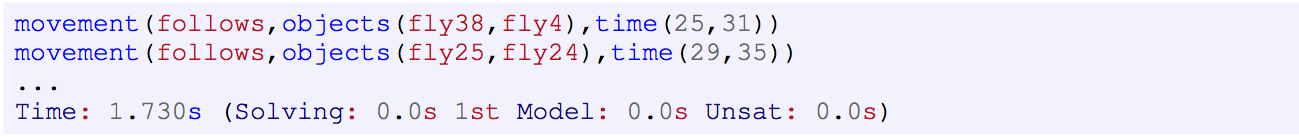}

%\scriptsize
%\begin{minted}[bgcolor=blue!5!white]{prolog}
%movement(follows,objects(fly38,fly4),time(25,31))
%movement(follows,objects(fly25,fly24),time(29,35))
%...
%Time: 1.730s (Solving: 0.0s 1st Model: 0.0s Unsat: 0.0s)
%\end{minted}
%\normalsize

\smallskip

The extract of the results shows that: $fly38$ follows $fly4$ during time $[25, 31]$;  $fly25$ follows $fly24$ during time $[29, 35]$.

\medskip

\exbul{2. CELL FUNCTION}\quad In this section we demonstrate how to solve spatial reasoning problems by translating polygons. Figure~\ref{fig:medical}(b) presents a stained tissue section of red and white blood cells from a patient with chronic myelogenous leukemia. We analyse the relationships between the physical structures of cell components, in particular whether certain cell components could move and fit inside other cell components. We segment the image, which assigns a class type to each segment, and apply standard contour detection algorithms to convert the raster image into polygons. We then parse the output as ASP facts including \emph{st\_object/3} and \emph{polygon/2}.

%  \todo{Need to change this to a spatio-temporal example, also ideally 3D - I'm considering a 3D dataset where the task is to recognise cell split events, this will work nicely with split/merge relation we define earlier} 

\smallskip

\noindent\textbf{Example 2.1}.\quad  {\sffamily\small  Firstly we determine whether a cell with the same shape as ``\emph{co:8}'' might also fit inside the cytoplasm region by creating a new polygon ``\emph{tr:8}'' that is a \emph{translation} of polygon ``\emph{co:8}''. We translate ``\emph{tr:8}'' so that it is a \emph{proper part} (pp) of ``\emph{co:127}''. }

\includegraphics[width=\textwidth]{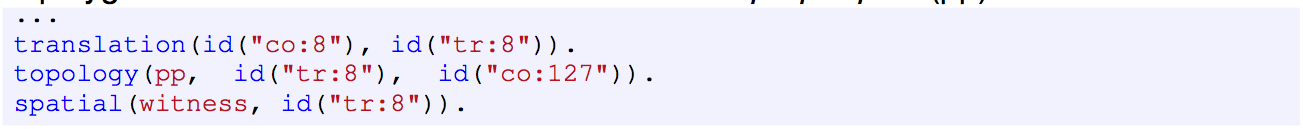}

%\scriptsize
%\begin{minted}[bgcolor=blue!5!white]{prolog}
%...
%translation(id("co:8"), id("tr:8")).
%topology(pp,  id("tr:8"),  id("co:127")).
%spatial(witness, id("tr:8")).
%\end{minted}
%\normalsize

\smallskip

\result

\smallskip

\includegraphics[width=\textwidth]{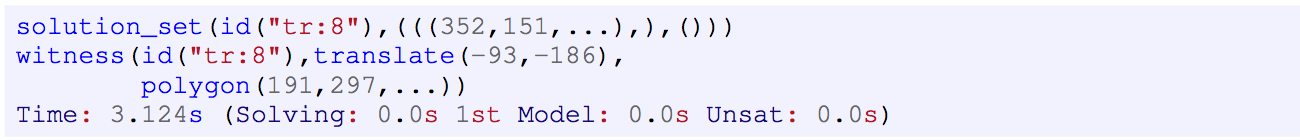}

%\scriptsize
%\begin{minted}[bgcolor=blue!5!white]{prolog}
%solution_set(id("tr:8"),(((352,151,...),),())) 
%witness(id("tr:8"),translate(-93,-186),
%        polygon(191,297,...))
%Time: 3.124s (Solving: 0.0s 1st Model: 0.0s Unsat: 0.0s)
%\end{minted}
%\normalsize

\smallskip

The result shows that indeed a cell with a polygon contour ``\emph{co:8}'' could be a proper part of the cytoplasm region with polygon contour ``\emph{co:8}'', and we are given a ground polygon as a witness that is a translation $t = (-93,-186)$ of polygon ``\emph{co:8}'' (by default, the witness given is the minimum translation required to satisfy the relation). 

\smallskip

\noindent\textbf{Example 2.2}.\quad  {\sffamily\small We now demonstrate going beyond purely qualitative reasoning by taking polygon \emph{shape} into account. We check whether ``\emph{tr:8}'' can be disconnected from \emph{both} ``\emph{co:139}'' and ``\emph{co:140}'' simultaneously (which is impossible due to the particular polygons in the dataset).}

\smallskip

\includegraphics[width=\textwidth]{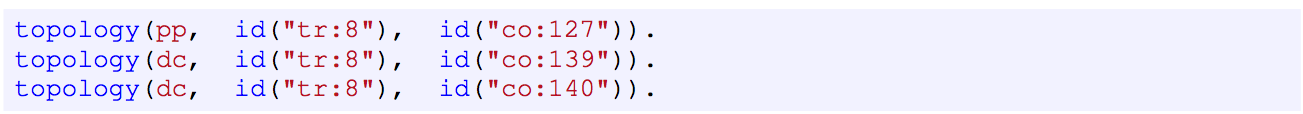}

%\scriptsize
%\begin{minted}[bgcolor=blue!5!white]{prolog}
%topology(pp,  id("tr:8"),  id("co:127")).
%topology(dc,  id("tr:8"),  id("co:139")).
%topology(dc,  id("tr:8"),  id("co:140")).
%\end{minted}
%\normalsize

\smallskip

\result

\smallskip

\includegraphics[width=\textwidth]{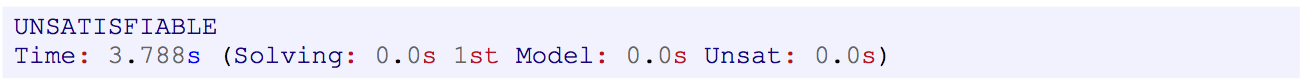}

%\scriptsize
%\begin{minted}[bgcolor=blue!5!white]{prolog}
%UNSATISFIABLE
%Time: 3.788s (Solving: 0.0s 1st Model: 0.0s Unsat: 0.0s)
%\end{minted}
%\normalsize

\smallskip

The result shows that no translation of polygon ``\emph{co:8}'' exists that satisfies all given topological constraints, due to the shapes of the polygons, i.e. this is an example of mixed qualitative-numerical reasoning.

%\subsection{Motion Planning by Spatio-Temporal Abduction}

\medskip

\exbul{3. MOTION PLANNING}\quad We show how \ST regions can be used for motion planning, e.g. in robotic manipulation tasks using abduction. 

\smallskip

\noindent\textbf{Example 3}.\quad  {\sffamily\small  An agent (a robot with a manipulator) is at a desk in front of a laptop. A cup of coffee is positioned behind the laptop and the agent wants to get the cup of coffee without the risk of spilling the coffee on the laptop. The agent should not hit the computer while performing the task.}

This task requires abducing intermediate states that are consistent with the domain constraints. We model the laptop, hand, and cup from a top-down perspective as \ST regions with polygonal slices, and give the initial shapes.

%The goal is given by the final position of the cup.

\smallskip

\includegraphics[width=\textwidth]{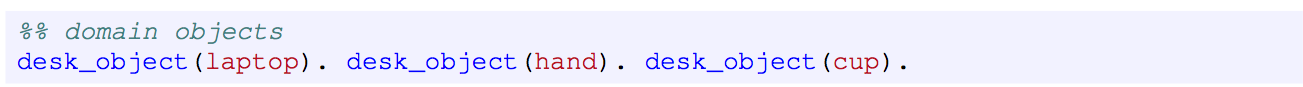}

%\scriptsize
%\begin{minted}[bgcolor=blue!5!white]{prolog}
%%% domain objects
%desk_object(laptop). desk_object(hand). desk_object(cup).
%\end{minted}
%\normalsize

\smallskip

The initial configuration is given for time $0$: 

\smallskip

\includegraphics[width=\textwidth]{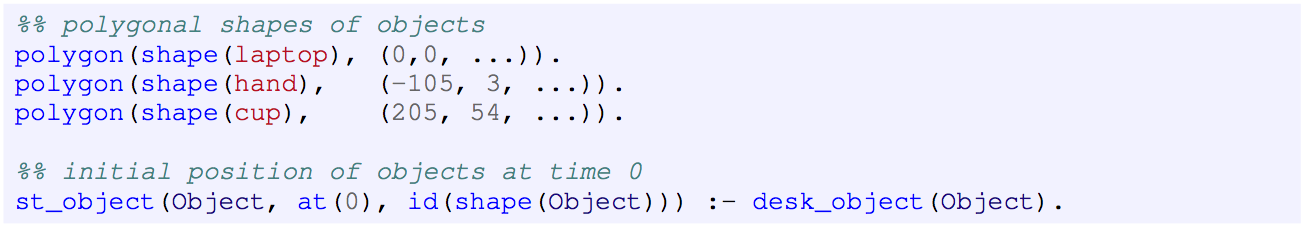}

%\scriptsize
%\begin{minted}[bgcolor=blue!5!white]{prolog}
%%% polygonal shapes of objects
%polygon(shape(laptop), (0,0, ...)).  
%polygon(shape(hand),   (-105, 3, ...)).
%polygon(shape(cup),    (205, 54, ...)).
%
%%% initial position of objects at time 0
%st_object(Object, at(0), id(shape(Object))) :- desk_object(Object).
%\end{minted}
%\normalsize

\smallskip

We model the scenario from time $0$ to $2$.

\includegraphics[width=\textwidth]{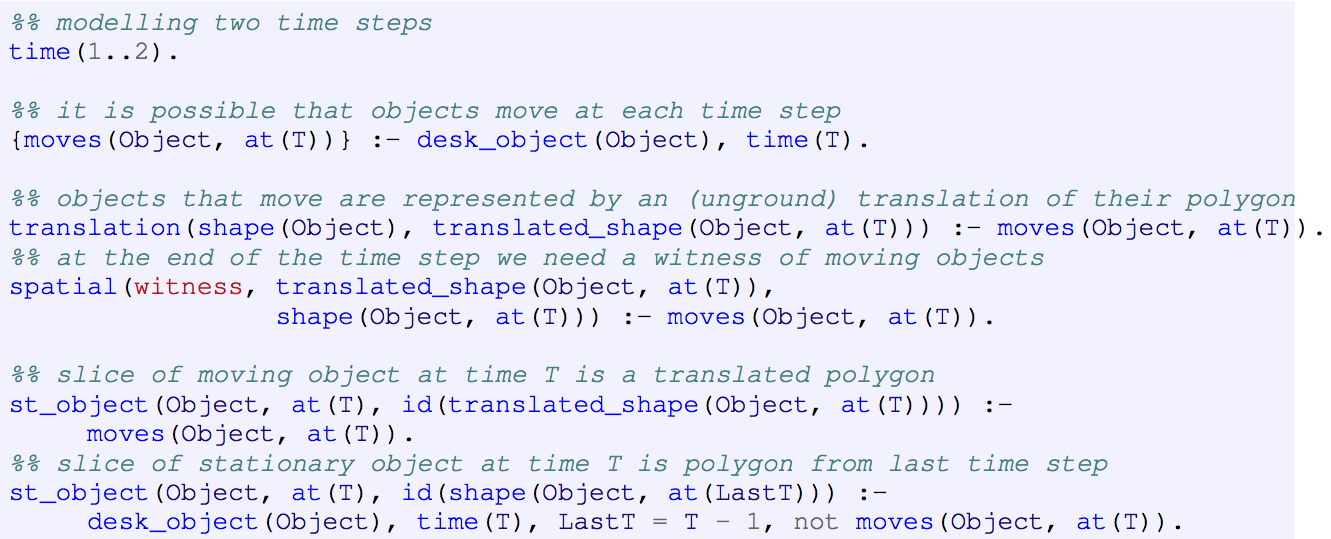}

%\scriptsize
%\begin{minted}[bgcolor=blue!5!white]{prolog}
%%% modelling two time steps
%time(1..2).
%
%%% it is possible that objects move at each time step
%{moves(Object, at(T))} :- desk_object(Object), time(T).
%
%%% objects that move are represented by an (unground) translation of their polygon 
%translation(shape(Object), translated_shape(Object, at(T))) :- moves(Object, at(T)). 
%%% at the end of the time step we need a witness of moving objects
%spatial(witness, translated_shape(Object, at(T)),
%                 shape(Object, at(T))) :- moves(Object, at(T)). 
%
%%% slice of moving object at time T is a translated polygon
%st_object(Object, at(T), id(translated_shape(Object, at(T)))) :-
%     moves(Object, at(T)).  
%%% slice of stationary object at time T is polygon from last time step
%st_object(Object, at(T), id(shape(Object, at(LastT))) :-
%     desk_object(Object), time(T), LastT = T - 1, not moves(Object, at(T)).  
%\end{minted}
%\normalsize

\smallskip

The goal is for the hand to make contact with the cup:

\includegraphics[width=\textwidth]{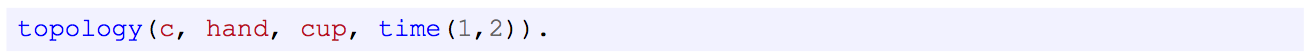}

%\scriptsize
%\begin{minted}[bgcolor=blue!5!white]{prolog}
%topology(c, hand, cup, time(1,2)).
%\end{minted}
%\normalsize

\smallskip

We model default domain assumptions, e.g., the cup does not move by default. We express this by assigning costs to interpretations where objects move.

\smallskip

\includegraphics[width=\textwidth]{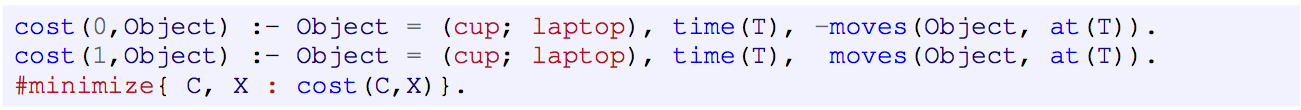}

%\scriptsize
%\begin{minted}[bgcolor=blue!5!white]{prolog}
%cost(0,Object) :- Object = (cup; laptop), time(T), -moves(Object, at(T)).
%cost(1,Object) :- Object = (cup; laptop), time(T),  moves(Object, at(T)).
%#minimize{ C, X : cost(C,X)}.
%\end{minted}
%\normalsize

\smallskip

The spatio-temporal constraints for planning the motion trajectory are that the hand and cup must remain disconnected from the laptop.

\smallskip

\includegraphics[width=\textwidth]{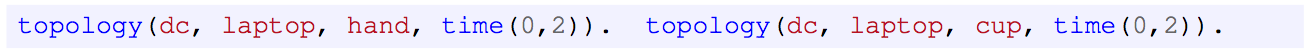}

%\scriptsize
%\begin{minted}[bgcolor=blue!5!white]{prolog}
%topology(dc, laptop, hand, time(0,2)).  topology(dc, laptop, cup, time(0,2)).
%\end{minted}
%\normalsize

\smallskip

Our system finds a consistent and optimal answer set where neither the laptop nor cup move in the period before the robot hand has made contact with the cup. Given the spatio-temporal constraints in this optimal answer set, our system then produces a consistent motion trajectory witness of the solution set (Fig. \ref{fig:medical}).

\subsection{Empirical Evaluation}
In the previous section we demonstrated applicability and runtime results of our system on real world data. We now empirically evaluate our system on synthetic data to more precisely assess runtime scalability and robustness against missing data in the following tests $T1 - T4$.\footnote{Experiments were run on a MacBook Pro, OSX 10.8 2.6 GHz, Intel Core i7, 16GB RAM. Runtime results are ASP grounding time plus solving time, as reported by clingo.}

\exbul{T1 (scalability / qualification)}\quad  Measuring runtime of deriving spacetime relations between $n$ \ST objects over $m$ time steps (Table~\ref{tab:t1}). Each ST object is assigned a randomly generated polygon slice (with between $5$ and $10$ vertices) for each time step. Each object has a direction vector, speed, fixed angular speed, and fixed acceleration (fixed values randomly selected from $[-0.1, 0.1]$). At each time step the object position is updated according to the direction and speed, and the direction and speed are updated according to the angular speed and acceleration. It is useful to identify semantically relevant object pairs based on other spatio-temporal relations, e.g. with social flies (Fig. \ref{fig:medical}) the \emph{follow} event is only meaningful when the flies are \emph{near}. We therefore measure (a) average time to compute relations between one pair of \ST objects for all time steps, (b) average time to compute relations between all \ST objects for one time step. Results show that our approach is practical within $n=40$ \ST objects and $m=40$ timesteps.

\begin{table}[t]
\footnotesize
\begin{center}
\caption{{\sffamily \footnotesize T1: Average runtime (seconds) for deriving spacetime relations.}}
\label{tab:t1}
\begin{tabular}{|l|r|r|r|r|}
\hline
%time: \textbackslash ~ objects:& $10$ & $20$ & $30$ & $40$ \\
$n$ \ST objects & $10$ & $20$ & $30$ & $40$ \\
(time steps $m=40$)& & & & \\
\hline
%One pair, all timesteps (sec)& $0.47$ & $1.11$ & $2.08$ & $3.3$\\
%All pairs, one timesteps (sec)& $0.46$ & $8.12$ & $7.72$ & $18.29$\\
One pair, all timesteps& $0.47$s & $1.11$s & $2.08$s & $3.3$s\\
All pairs, one timestep& $0.46$s & $8.12$s & $7.72$s & $18.29$s\\
\hline
\hline
$m$ time steps & $10$ & $20$ & $30$ & $40$ \\
(\ST objects $n=40$)& & & & \\
\hline
One pair, all timesteps & $0.33$s & $0.92$s & $1.88$s & $3.24$s\\
All pairs, one timestep & $15.47$s & $15.89$s & $16.87$s & $18.14$s\\
\hline
\end{tabular}
\end{center}
\end{table}

\begin{table}
\footnotesize
\begin{center}
\caption{{\sffamily \footnotesize T2: Accuracy of derived relations from interpolation when $k$ slices are deleted ($k \in \{10,20,30,40\}$) from $200$ slices.}}
\label{tab:t2}
\begin{tabular}{|l|r|r|r|r|}
\hline
%time: \textbackslash ~ objects:& $10$ & $20$ & $30$ & $40$ \\
Deleted slices: & $5\%$ & $10\%$ & $15\%$ & $20\%$ \\
\hline
Correct relations& $97.32\%$ & $95.90\%$ & $94.41\%$ & $93.02\%$ \\
\hline
\end{tabular}
\end{center}
\end{table}

\begin{table}
\footnotesize
\begin{center}
%\caption{{\sffamily \footnotesize Test T4: Runtime (seconds) for finding all models determining (in)consistencies in finding polygon translations to satisfy qualitative constraints.}
%\caption{{\sffamily \footnotesize Test T4: Runtime (seconds) for finding all models of combinations of qualitative constraints.}
\caption{{\sffamily \footnotesize T3: translating $m=10$ polygon slices of \ST object $g$ to satisfy qualitative constraints (find first $10,000$ models).}
}
\label{tab:t4}
\begin{tabular}{|l|r|r|r|r|}
\hline
%time: \textbackslash ~ objects:& $10$ & $20$ & $30$ & $40$ \\
$n$ ST objects: & $5$ & $10$ & $15$ & $20$ \\
\hline
Runtime (sec) & $1.30$s & $6.15$s & $17.55$s & $39.38$s\\
\hline
\end{tabular}
\end{center}
\end{table}

\begin{table}[t]
\footnotesize
\begin{center}
\caption{{\sffamily \footnotesize T4: Runtime (seconds) for determining inconsistencies in purely qualitative constraints.}
}
\label{tab:t3}
\begin{tabular}{|l|r|r|r|r|}
\hline
%time: \textbackslash ~ objects:& $10$ & $20$ & $30$ & $40$ \\
$n$ ST objects: & $10$ & $20$ & $30$ & $40$ \\
\hline
Models (mean)& $20\%$ & $10\%$ & $10\%$ & $20\%$\\
Runtime (sec)& $0.8562$ & $3.7995$ & $10.7358$ & $205.941$\\
\hline
\end{tabular}
\end{center}
\end{table}

\exbul{T2 (robustness)}\quad  Measuring accuracy of derived spacetime relations when slices are randomly deleted from \ST objects (Table~\ref{tab:t2}). Tests are created as in T1 with $10$ objects over $20$ time steps. In each such test $t$ there are $m \times n$ polygon slices. We copy $t$ to create test $t^\prime$, randomly select $k$ slices and delete them from $t^\prime$. We then compare \ST relations derived from $t$ and $t^\prime$ and record the number of matching relations as a measure of accuracy. Our results indicate that linearly interpolating between slices is satisfactorily robust against missing data. This also implies that using ASP to sample large datasets to reduce the search space when identifying meaningful spatio-temporal relations is a viable approach.

\medskip

\exbul{T3 (scalability / translation)}\quad  Measuring runtime of determining (in)consistency of translating a polygon to satisfy given spacetime constraints (mixed-numerical reasoning problem) (Table~\ref{tab:t4}). For each test, $n$ \ST objects are created as in T1, and a new \ST object $g$ is declared and assigned $m=10$ randomly generated polygon slices that can be translated. We measure time taken to find the first $10,000$ models (and solution sets of all consistent translations) where one mereotopological relation is asserted between $g$ and each other object (i.e. each model has $n$ relations). The large number of models is due to existential \ST relations, e.g. two \ST objects have \emph{contact} if \emph{at least one slice} has contact, thus leading to many alternative models. The results show that our approach is practical up to $n=20$ objects.

\medskip

\exbul{T4 (scalability / inconsistency)}\quad  Measuring runtime for determining (in)consistency of $n$ qualitatively constrained \ST objects with no numerical information (purely qualitative reasoning) (Table~\ref{tab:t3}). Each object $i \in \{1,2,\ldots,n\}$ is declared with no polygonal slices. Object pairs are randomly selected and assigned $4$ randomly chosen alternative \ST relations using the algorithm described in \cite{renz2001efficient} (mean degree of constraint network $d=5$). Each test with $n$ objects is run $10$ times, we report mean runtime and number of models (i.e. consistent constraint networks). Our results show that our approach is practical up to $n=30$ objects before combinatorial explosion occurs.

\section{\uppercase{Discussion and Related Work}}

ASP Modulo extensions for handling specialised domains and abstraction mechanisms provides a powerful means for the utilising ASP as a foundational knowledge representation and reasoning (KR) method for a wide-range of application contexts. This approach is clearly demonstrated in work such as ASPMT \citep{joohyung-aspmt-2013,bartholomew2014system,gelfond2008answer}, \textsc{Clingcon} \cite{geossc09a}, ASPMT(QS) \citep{ASPMTQS-LPNMR-2015}. Most closely related to our research is the ASPMT founded \emph{non-monotonic spatial reasoning} system  \textsc{ASPMT(\QS)} \cite{ASPMTQS-LPNMR-2015}. Whereas \textsc{ASPMT(\QS)} provides a valuable blueprint for the integration and formulation of geometric and spatial reasoning within answer set programming modulo theories, the developed system is a first-step and lacks support for a rich spatio-temporal ontology or an elaborate characterisation of complex `space-time' objects as native (the focus there has been on enabling non-monotonicity with a basic spatial and temporal ontology). In addition to the ontological extensions for a much richer `space-time' component, our system pipeline --based on \textsc{clingo} \citep{gekakasc14b} --- has the following additional advantages over the standard ASPMT / ASPMT(\QS) pipeline:\quad  \textbf{(1)}. we generate \emph{all} spatially consistent models compared to only one model in the standard ASPMT pipeline;\quad \textbf{(2)}.  we compute optimal answer sets, e.g. add support preferences, which allows us to rank models, specify weak constraints;\quad \textbf{(3)}.  unlike ASPMT(\QS) we support quantification of space-time regions.

%\subsubsection{\uppercase{Foundations -- The RCC8 Language}}
\medskip

Within the relation algebraic driven (qualitative) spatial reasoning community, researchers have investigated translating qualitative spatial calculi into ASP programs e.g. \cite{li2012qualitative,brenton_et_al:OASIcs:2016:6735}. The primary difference with our line of research is we emphasise both purely qualitative and mixed qualitative-quantitative constraints and efficiently deriving \ST relations from large datasets, and that spato-temporal entities and relations have natively encoded semantics within the KR framwork being employed, namely answer set programming. More broadly, this research is driven by a departure from the use of relational-algebra, and instead focussing on \emph{declarative spatial reasoning} directly within KR frameworks such as constraint logic programming, answer set programming, and inductive logic programming \citep{DBLP:conf/cosit/BhattLS11,ASPMTQS-LPNMR-2015,DBLP:conf/ilp/SuchanBS16}.

\section{\uppercase{Summary and Outlook}}

A novel method and corresponding system for declaratively modelling and reasoning about the dynamics of space-time histories ---\emph{regions with spatial and temporal components}--- as first-class objects within answer set programming is developed. The framework is implemented as an extension of the \textsc{Clingo} ASP solver \citep{gekakasc14b}, whereas the crux of the method relies on leveraging upon the semantics of (mereotopological) spatio-temporal relations using specialised and highly optimised systems of polynomials. We have presented an empirical evaluation, and  demonstrated several reasoning features in the context of select applications domains requiring interpretation and control tasks. The outlook of this work is geared towards enhancing the application of the developed specialised ASP Modulo Space-Time component specifically for non-monotonic spatio-temporal reasoning about large datasets in the domain of visual stimulus interpretation, as well as constraint-based motion control in the domain of home-based and industrial robotics. The reasoning system is also slated for deployment as an open-source robotics domain specific library as part of the ROS \cite{quigleyros} robotics framework.

\footnotesize
\bibliographystyle{splncs04}
\small
\bibliography{ijcai18,space-time-motion,QSR-bib}

\end{document}